\documentclass[conference]{IEEEtran}
\IEEEoverridecommandlockouts

\usepackage{cite}
\usepackage{amsmath,amssymb,amsfonts}
\usepackage{algorithmic}
\usepackage{graphicx}
\usepackage{booktabs}
\usepackage{hyperref}
\usepackage{array}
\usepackage{graphicx}
\usepackage{textcomp}
\usepackage{subcaption}
\usepackage{multicol}
\usepackage{multirow}
\usepackage{xcolor}
\usepackage{lscape}
\def\BibTeX{{\rm B\kern-.05em{\sc i\kern-.025em b}\kern-.08em
    T\kern-.1667em\lower.7ex\hbox{E}\kern-.125emX}}
\begin{document}

\title{CRAVE: A Conflicting Reasoning Approach for Explainable Claim Verification Using LLMs
}

\author{

	\IEEEauthorblockN{
		Yingming Zheng \textsuperscript{a},
		Xiaoliang Liu \textsuperscript{c},
            Peng Wu \textsuperscript{a,b}*\thanks{*Corresponding author.},
            and Li Pan \textsuperscript{a,b}
            }
	
        \IEEEauthorblockA{\textsuperscript{a}School of Computer Science, Shanghai Jiao Tong University, Shanghai, China}
	\IEEEauthorblockA{\textsuperscript{b}Shanghai KEY Laboratory of Integrated Administration Technologies for Information Security, Shanghai, China}
    \IEEEauthorblockA{\textsuperscript{c}School of Politics, National Defense University, Shanghai, China\\
        zhengyingming@sjtu.edu.cn,lxltqz@sjtu.edu.cn, catking@sjtu.edu.cn, panli@sjtu.edu.cn}
              
}

\maketitle

\begin{abstract}
The rapid spread of misinformation, driven by digital media and AI-generated content, has made automatic claim verification essential. Traditional methods, which depend on expert-annotated evidence, are labor-intensive and not scalable. Although recent automated systems have improved, they still struggle with complex claims that require nuanced reasoning. To address this, we propose CRAVE, a Conflicting Reasoning Approach for explainable claim VErification, that verify the complex claims based on the conflicting rationales reasoned by large language models (LLMs). Specifically, CRAVE introduces a three-module framework. Ambiguity Elimination enchanced Evidence Retrieval module performs ambiguity elimination and entity-based search to gather relevant evidence related to claim verification from external sources like Wikipedia. Conflicting Perspective Reasoning and Preliminary Judgment module with LLMs adopts LLMs to reason rationales with conflicting stances about claim verification from retrieved evidence across four dimensions, i.e., direct evidence, semantic relationships, linguistic patterns, and logical reasoning and make a preliminary judgment. Finally, Small Language Model (SLM) based Judge module is fine-tuned to make use of preliminary judgment from LLMs to assess the confidence of the conflicting rationales and make a final authenticity judgment. This methodology allows CRAVE to capture subtle inconsistencies in complex claims, improving both the accuracy and transparency of claim verification. Extensive experiments on two public claim verification datasets demonstrate that our CRAVE model achieves much better performance than state-of-the-art methods and exhibits a superior capacity for finding relevant evidence and explaining the model predictions. The code is
provided at \url{https://github.com/8zym/CRAVE}.
\end{abstract}


\section{INTRODUCTION}
The rapid development of digital media and the internet, along with the continuous improvement of AIGC technologies, has led to a rise in the spread of misinformation, which poses significant risks to political, economic and other fields\cite{damage2}. To address this, automatic claim verification systems\cite{automatedfactchecking}, designed to detect and identify misleading or inaccurate information online, have become increasingly crucial.\par 
Early automatic claim verification work relied on experts with domain-specific knowledge to manually annotate golden evidence for each claim. The truthfulness of the claim was determined based on whether golden evidence supported or refuted the claim. Annotating golden evidence for each claim is labor-intensive, making it difficult to scale for processing large volumes of claims in real-world applications. Recently, to avoid manual annotation of evidence, many claim verification models\cite{pipeline4} use an automated pipeline consisting of two steps: evidence retrieval and verdict prediction. These methods eliminate the need for pre-annotated evidence by automatically retrieving evidence from external sources to support the truthfulness judgment of the claim. Typically, these methods retrieve evidence from external source, such as Wikipedia, based on its relevance to the claim, and then use neural network models to capture the consistency or contradiction between the evidence and the claim to assess the truthfulness of the claim. These methods have achieved promising results in verifying the truthfulness of simple claims.\par 
However, with the maturity of generative artificial intelligence, such as large language models (LLMs), more and more complex and suspicious claims are emerging. They are usually logically complex and involve information from multiple aspects. The method of directly retrieving evidence based on its relevance to the claims may be difficult to obtain comprehensive evidence required for verifying authenticity. Moreover, neural network based verification models \cite{neturalnetwork}  may be hard to capture subtle consistency relationships between complex claims and evidence to judge the authenticity, and provide explainable justifications. \par 

In recent years, the powerful semantic understanding and clue-sensitivity capabilities of LLMs have shown potential to better understand logically complex claims and identify subtle cues within evidence that support or refute the claims\cite{LLMpotential2,LLMpotential3}, thereby enabling more accurate truthfulness assessment. Previous studies\cite{LLMdirect2} have directly utilized LLMs to analyze evidence for supporting or refuting claims without any specific instructions. Compared to allowing LLMs to reason freely without direction, providing a specific reasoning direction\cite{GoodActor} may avoid LLMs being confused by opposing viewpoints and enable LLMs to more fully explore the cues relevant to that direction. Recent works\cite{Harmfulmeme,LLMdebate1,LLMdebate2} have begun to guide LLMs to analyze evidence from both the supporting and refuting perspectives and design neural networks for information integration and judgment. 
However, most of them have not fully leveraged the characteristics of misinformation, which differ semantically, in linguistic patterns, and in commonsense reasoning compared to truth.\par 
Therefore, we propose CRAVE, a \textbf{C}onflicting \textbf{R}easoning \textbf{A}pproach for for explainable claim \textbf{VE}rification approach, which adopts LLMs to reason the rationales for verifying the authenticity of claims from the conflicting perspectives of both true and fake. Specifically, to obtain the corresponding evidence, CRAVE introduces a Ambiguity Elimination enchanced Evidence Retrieval module that performs ambiguity elimination and entity extraction for the most pertinent evidence retrieval. 
Based on the retrieved evidence, CRAVE incorporates a Conflicting Perspective Reasoning and Preliminary Judgment module with LLMs, which utilizes an LLM to reason rationales about claim verification from the evidence from two conflicting perspectives, i.e., whether the evidence supports the claim or refutes it, and make a preliminary prediction based on the reasoned rationales. The rationales of each perspective include four dimensions: direct evidence analysis, semantic features and relationships, linguistic patterns and connections, and logical reasoning\cite{fouraspect2,fouraspects}. Subsequently, based on the generated rationales from conflicting perspectives and preliminary prediction, CRAVE fine-tunes a Small Language Model (SLM) based Judge module to automatically capture the confidence of both perspectives and make the final authenticity judgment. Through the above steps, we can infer the final verification result from the extracted evidence, generate justifications, and return the relevant evidence along with its corresponding URLs, aiding readers in making a reasonable judgment about the validity of the result. 
Our contributions are summarized as follows:
\begin{itemize}
    \item We highlight that the conflicting perspective reasoning is important for LLMs to avoid being confused by opposing viewpoints and more fully explore the cues relevant claim verification.
    \item We propose CRAVE, which retrieves the most pertinent evidence based on ambiguity elimination and entity extraction, then adopts LLMs to reason about rationales for claim verification from two conflicting perspectives as well as generate a preliminary judgment, and finally designs a SLM based judgment module to verify claims.
    \item We conducted experiments on two real-world datasets, and the results show that our model achieves state-of-the-art performance while providing robust explanations.
\end{itemize}

\section{RELATED WORK}
\subsection{Claim Verification}
With the development of deep learning technologies, many studies have proposed automated claim verification models. These models typically consist of three main steps: 1) Claim decomposition: breaking down complex claims into smaller sub-claims; 2) Evidence retrieval: searching for documents and sentences related to the claim from structured or unstructured data; 3) Verify prediction: using classification models to compare the claim with the evidence and determine its truthfulness. 
The Factify-5WQA\cite{5WQA} model proposed by Anku Rani et al. decomposes the claim into five possible W-questions and retrieves corresponding evidence for each question, improving the logical consistency of claim decomposition. Jiho Kim et al.\cite{FactKG} extract triples from claims using DBpedia, categorize them into five distinct types, and perform evidence retrieval by reasoning over the knowledge graph. These methods optimize sentence decomposition and sub-claim verification to obtain the final answer. Hao Liao et al.'s MUSER model\cite{MUSER} moves beyond the three-step framework and simulates human reasoning processes by proposing a multi-step evidence extraction approach. While these methods have made significant strides in improving the accuracy of evidence extraction and subclaim verification, there are still opportunities for further enhancement, particularly in terms of reasoning over the evidence and providing comprehensive, high-quality justifications for the final conclusions. \par 

\subsection{LLMs based methods}
In recent years, LLMs have made significant progress in claim verification, as they leverage their deep language understanding and reasoning abilities, enabling them to handle complex reasoning tasks. By generating reasoning processes and explanations through Chain-of-Thought (CoT) prompting, LLMs improve accuracy in multi-hop claim verification. 
Pan et al. \cite{ProgramFC}utilize LLMs to divide the claim into three distinct tasks, process each task with different sub-models, and combine the results logically to derive the final outcome. The FOLK model proposed by Haoran Wang and Kai Shu\cite{FOLK} applies First-Order Logic to decompose the problem through LLMs, uses the Google API to expand the scope of evidence retrieval, and employs LLMs to generate explanations and make final decision. Though LLMs are excellent at reasoning and in-context learning, they face challenges: LLMs are less reliable in judgment tasks compared to fine-tuned models like BERT, and may generate hallucinated information that appears logical but is factually incorrect\cite{hallucinate1,hallucinate3}. \par 
Recent studies have leveraged LLMs to uncover deep information from text and images and adopt SLM to make judgment based on deep information. For instance, Hu et al.\cite{GoodActor} analyzed fake news' textual and semantic features, while Lin et al.\cite{Harmfulmeme} explored potential harm in comics. Inspired by these insights, we use LLMs to resolve ambiguous references and extract entities through external evidence retrieval, generate true/false rationales to support explainable claim verification and make a preliminary judgment to assess confidence of two sides, enhancing the fine-tuning of smaller language models.



\section{OUR APPROACH}

\begin{figure*}[t]
    \centering
    \includegraphics[width=0.9\textwidth]{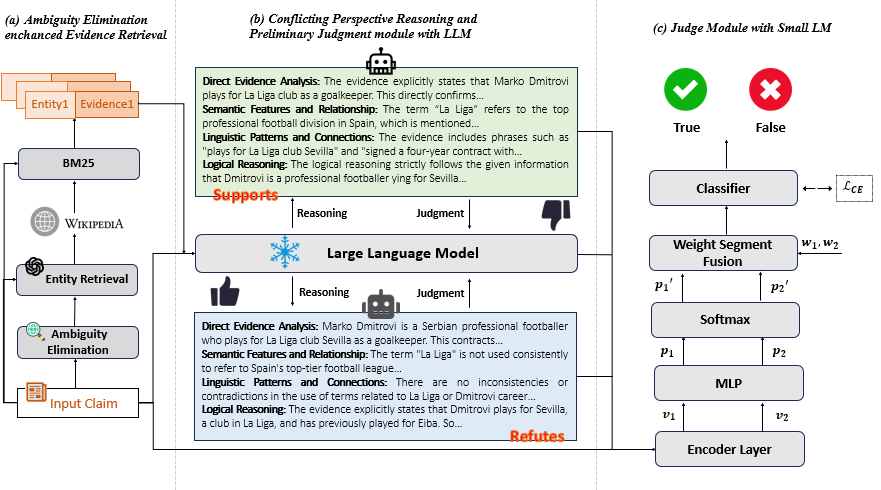}
    \caption{The overview of CRAVE. It consists of three core module. a) Ambiguity Elimination enchanced Evidence Retrieval Module contains ambiguity elimination, entity retrieval and evidence retrieval and selection parts. b) Conflicting Perspective Reasoning and Preliminary Judgment module with LLMs that reason why the evidence supports or refutes the claim from four reasoning aspects and make a preliminary Judgment using LLMs. c) Judge module adopts SLM to make the final decision.}
    \label{fig:model}
\end{figure*}

\subsection{Problem Statement}
Given a claim $C$ and corresponding knowledge source $K$, we extract relevant evidence from the knowledge source based on the claim, obtaining a set of evidence paragraphs $[d_1,\dots,d_n]$ (where $d_i\in K, i\in [1,n])$, donated as $D$. Then we use a model $F$ to reason over the evidence and generate the final explanation. Therefore, the problem can be defined as follows: $F(C,K) = (\hat{y},e)$, $\hat{y}$ represents authenticity labels and e represents generated explanation.\par 
In our model, we first perform ambiguity elimination on claim $C$ to extract entities, resulting in $E=[e_1,\dots,e_n]$. For each entity $e_i$, we retrieve the most relevant evidence $D_i$ from external source using BM25 and gather it into $D$. Using LLMs, we extract clues $r_{true},r_{false}$ from the evidence set $D$ and make a preliminary prediction $y_{LLM}$. Then we combine clues with the original claim in a certain sequence based on $y_{LLM}$ to make the final judgment and explanation through SLM.\par 
The specific framework of the model is shown in Figure \ref{fig:model}. It consists of the following components: ambiguity elimination enchanced Evidence Retrieval Module(Section \ref{sec:3.2}), Conflicting Perspective Reasoning and Preliminary Judgment Module with LLMs(Section \ref{sec:3.3}), and the Judge Module with Small LM (Section \ref{sec:3.4}).

\subsection{Ambiguity Elimination enchanced Evidence Retrieval Module}\label{sec:3.2}
\subsubsection{Ambiguity Elimination}

\begin{figure}[t]
    \centering
    \includegraphics[width=0.5\textwidth]{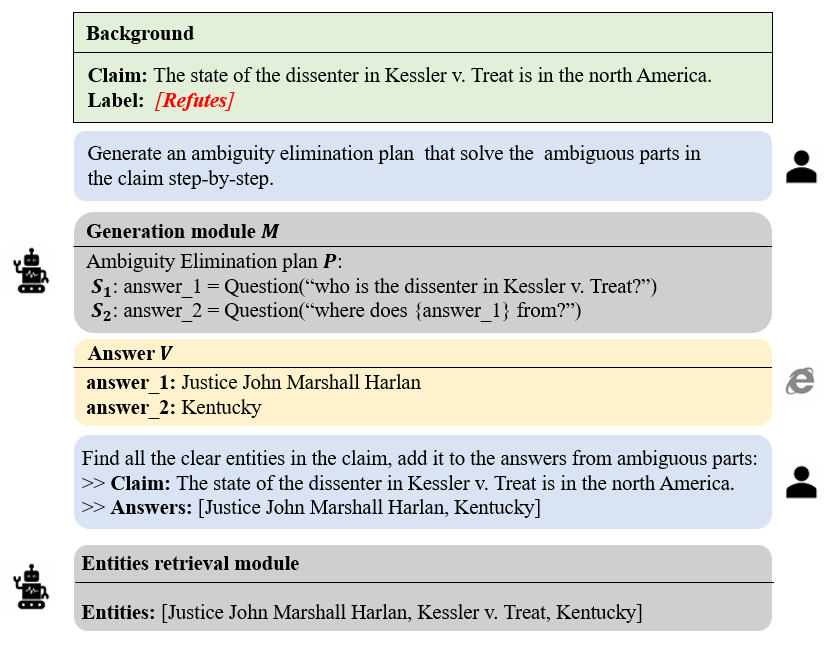}
    \caption{Simple example for the ambiguity elimination and entity retrieval in Ambiguity Elimination enchanced Evidence Retrieval Module. After these procedures, CRAVE will find evidence based on the entities.}
    \label{fig:Example}
\end{figure}

Many claims contain ambiguous entities that need to be indirectly determined through contextual descriptions\cite{ambi}. When retrieving evidence, relying solely on explicit entities mentioned in the claim may overlook evidence related to ambiguous entities that are crucial for claim verification. Therefore, we first perform ambiguity elimination to identify specific and well-defined entities for all ambiguous entities in the claim. Although models like BERT and RoBERTa excel in many NLP tasks, their effectiveness relies heavily on carefully curated datasets and well-designed training processes to achieve optimal performance. Thus we employ LLMs for ambiguity elimination leveraging their powerful semantic capabilities. \par 
We give a simple example in Figure \ref{fig:Example}. Given the input claim $C$, a generation module $M$ will use LLMs to generate an ambiguity elimination plan $P= [S_1,\dots,S_n]$, which consists of n sequentially ordered reasoning steps $S_i$. Each reasoning step $S_i \in P$ is displayed in natural language that represents an exact semantic ambiguity part in the claim. We define $S_i: A_i = Question(Q_i)$, where $Q_i$ is a question aiming to clarify ambiguous entities, and $A_i$  is the answer to the question. It should be noticed that $Q_i$ probably contains information which retrieved from previous steps for multi-hop claims. Therefore, the steps must be executed in a strict sequence.

\subsubsection{Entity Retrieval}
Retrieving relevant information based on key entities in a claim is a commonly used and effective information retrieval method. We prompt the LLMs to extract all unambiguous entities from the claim and combine them with the entities returned by the ambiguity elimination module to form a comprehensive entity set for evidence retrieval.  

\subsubsection{Evidence Retrieval and Selection }
Based on the entities returned in the previous step, we locate the corresponding external source page for each entity. The corresponding page of each entity contains comprehensive information about the entity. The first few paragraphs, which provide a summary of the entity, are more likely to be useful for verifying the claim's authenticity. Additionally, to avoid missing important information from other sections of the page, we further employ BM25 to retrieve the most relevant paragraphs from other parts of the page for claim verification. \par 
Specifically, for each entity $e_i$, we retrieve its corresponding external source page which is in the form of a set of paragraphs $[p_1,\dots,p_n]$. We retrieve the first two paragraphs as the entry's summary. Then, using the BM25 algorithm, we further rank and retrieve the most relevant paragraphs from the remaining content. After these, we obtain evidence blocks $D_i = [d_{i1},\dots,d_{im}]$ related to each entity $e_i$. By consolidating the evidence blocks $[D_1,\dots,D_i]$ corresponding to all entities, we gather all the evidence $D$ relevant to the claim. This process ensures that distractions are minimized while all critical reasoning evidence is extracted into $D$, facilitating subsequent reasoning by the LLM.

\subsection{Conflicting Perspective Reasoning and Preliminary Judgment Module with LLMs}\label{sec:3.3}
Although the retrieved evidence is potentially relevant to claim verification, it often contains substantial amounts of irrelevant and redundant information, which may reduce the accuracy and efficiency of claim verification and make it unsuitable as a direct explanation for verification results. Therefore, we employ an LLM to further analyze the retrieved evidence and extract the most critical justifications for claim verification. However, directly prompting the LLM to summarize key justifications may lead to ambiguous guidance, causing it to be confused by different perspectives present in the evidence. To address this, we provide explicit reasoning directions during LLM inference, guiding it to extract justifications from the evidence separately from opposing perspectives. Meanwhile, to standardize the LLM's output and ensure more comprehensive and reasonable answers, we perform rationale analysis from four aspects.\par 
Firstly, \textit{Direct Evidence Analysis} focuses on extracting parts of the evidence that can directly prove the truth or falsity of the claim. This includes specific data, facts, or information that clearly support or refute the claim. Secondly, \textit{Logical Reasoning} aims to analyze conclusions that can be logically derived from the evidence. This goes beyond explaining surface-level information in the evidence and requires using existing knowledge frameworks and logical rules to infer deeper connections. Thirdly, \textit{Semantic Features and Relationships} focuses on identifying and parsing the semantic features within the evidence and the relationships between them. By analyzing vocabulary and syntactic structures, we can uncover hidden semantic relationships that may significantly impact the claim’s verification. Finally, \textit{Linguistic Patterns and Connections} explores the language patterns and connections between texts in the evidence. This includes analyzing rhetorical devices, contrast relationships, and causal connectors, through which we can reveal implicit reasoning paths and logical connections between texts, further refining the judgment of the claim. \par 
Specifically, for claim $C$, designated label $y_p$ and corresponding evidence $D$, we design the following prompt $p^*$:\par 
\textit{"Analyze why the following claim [$C$] is [$y_p$] based on the given evidence [$E$]. Provide a clear and detailed explanation that focuses on: Direct evidence analysis; Semantic features and relationships; Linguistic patterns and connections; Logical reasoning strictly from the evidence."} \par 
The reasoning process is conducted separately for both the true and false scenarios, generating two distinct reasoning texts. It is observed that the rationale generated from the scenario consistent with the claim's truthfulness is typically more comprehensive and reasonable, outperforming the rationale extracted by competing scenario. Therefore, the confidence of the reasons presented for both stances differ. To aid the final judgment by the SLM, we first use the LLM to generate a preliminary judgment by assessing the confidence levels of the reasons supporting each stance. The prompt for this judgment is set as follows: \par 
\textit{"Given claim [$C$] and the following two claim rationales: (1) true:[$r_{true}$];(2) false:[$r_{false}$], is this claim true or false?"}\par 
Using this prompt, the LLM generates a preliminary prediction label $y_{LLM}\in\{true,false\}$, which provides its preference for one of the two labels, indicating which corresponding rationale is more reasonable. However, as previously discussed, due to the limitations of LLMs, we still require the SLM for the final judgment to ensure accuracy.

\subsection{Judge Module with Small LM}\label{sec:3.4}
We utilize a fine-tuned SLM to make the final judgment. The purpose of this SLM is to leverage the claim $C$ and the LLM-generated veracity-oriented rationales $r_{true},r_{false}$ as well as preliminary judgment $y_{LLM}$ to produce the final prediction label. As mentioned earlier, LLMs tend to generate strong reasoning to support true claims, while for false claims, they generate weaker reasoning, often containing less accurate information. Moreover, due to the inherent flaws in LLMs\cite{GoodActor}, we cannot fully trust the labels generated by the LLMs. Therefore, it is necessary to effectively combine the information from both the claim and rationales to aid the SLM in making the final inference. \par 

The claim $C$ is concatenated respectively with the reasons $r_{true}$ and $r_{flase}$ to form two text segments, which are then encoded separately using a Transformer encoder. Notably, in order for the SLM to capture the confidence levels of the two reasons as perceived by the LLM, the input sequence of the two reason segments is fixed in the model. The representation of the reason supporting the stance that aligns with the label predicted by the large model ($y_{LLM}$) is placed first, while the representation of the opposing reason is placed second:

\begin{equation}
\begin{aligned}
    v_1 &= \text{Transformer\_Enc}\left([C \; \text{[SEP]} \; r_{(a)} ; \theta^{\text{enc}}]\right) \\
    v_2 &= \text{Transformer\_Enc}\left([C \; \text{[SEP]} \; r_{(b)} ; \theta^{\text{enc}}]\right)
\end{aligned}
\end{equation}
where $(a),(b) \in \{true,false\}$, and (a) is equal to $y_{LLM}$ while (b) is contrast to $y_{LLM}$, showing that $v_1$ has higher confidence than $v_2$. In this case, SLM will pay more attention to $v_1$. Subsequently, we utilize a classifier to compute their respective prediction results.
\begin{equation}
\begin{aligned}
    p_1^{ver} &= \text{softmax}(\text{MLP}(v_1;\theta^{ver})) \\
    p_2^{ver} &= \text{softmax}(\text{MLP}(v_2;\theta^{ver}))
\end{aligned}
\end{equation}
Then we concatenate two predictions into final prediction:
\begin{equation}
    p^{ver} = w_1 \cdot p_1^{ver} + w_2 \cdot p_2^{ver}
\end{equation}
where $\theta^{enc},\theta^{ver},w_1,w_2$ are learnable parameters. $w_1,w_2$ are ranging from 0 to 1 and $w_1 + w_2 = 1$. The training objective of detection task can be represented as below:

\begin{equation}
    \mathcal{L} = \text{CE}(p^{ver},y) = -\sum_{D} logp^{ver}_{[y^{*} = y]}
\end{equation}
where CE represents the cross-entropy loss function, and $p^{ver}_{[y^{*} = y]}$ denotes the probability of predictions that match the true label. The final output can be simplified as:
\begin{equation}
    \hat{y} = \text{argmax} \; p^{ver}
\end{equation}
Based on the prediction, we can output relevant explanation:
\begin{equation}
    e = \left\{
\begin{array}{l}
    r_{true}, \hat{y} = true\\
    r_{false}, \hat{y} = false
\end{array}
\right.
\end{equation}
Since we use Wikipedia as knowledge source, in the final explanation output, we will include URLs of the relevant pages to allow users to review and discuss the results. This ensures transparency and helps mitigate the impact of misjudgments.

\section{EXPERIMENT}
\subsection{Dataset}
In our work, we focus on complex claims that need comprehensive understanding of evidence and claims or need multi-step reasoning. Given this context, we choose two challenging datasets: HOVER\cite{Hover} and FEVEROUS\cite{feverous}. Both datasets provide gold evidence for each claim.
 In our experiments, this gold evidence is used to evaluate all methods under gold-evidence setting to assess their fact verification capabilities. However, in real-world fact verification scenarios, gold evidence is not available in advance. Therefore, we additionally conducted experiments in an open-book setting, where evidence is retrieved from external knowledge sources for each claim, and its veracity is subsequently verified.

\subsubsection{HOVER}
HOVER contains claims that need multi-hop reasoning over knowledge source. We use the training dataset for train and validation dataset for test. The validation dataset contains 1126 two-hop claims, 1835 three-hop claims, and 1039 four-hop claims in total.
\subsubsection{FEVEROUS}
FEVEROUS is a dataset for complex claim verification over structured and unstructured data. As we focus on textual claim verification, we only use claims that require exclusively structured evidence, containing 2962 claims in total.

\subsection{Baseline}
We compare CRAVE against seven baselines, which can be categorized into three groups.
\subsubsection{Pretrained models} 
In this group, we use BERT-FC\cite{BERT-FC} and LisT5\cite{list5}, two models that leverage BERT and T5 for direct claim verification respectively. 
\subsubsection{NLI fine-tuned models}
In this group, we select three pretrained models fine-tuned on claim verification or natural language inference (NLI) datasets. RoBERTa-NLI\cite{Roberta-NLI} is a fine-tuned version of RoBERTa-large, trained on four NLI datasets. DeBERTaV3-NLI\cite{Deberta} is fine-tuned on 885,242 annotations from the FEVER dataset and four additional NLI datasets. MULTIVERS\cite{multivers}, a Longformer model, is fine-tuned on the FEVER dataset for improved performance in claim verification tasks. For both pretrained model and NLI fine-tuned model, in the open-book setting, we use BM25 implemented with the Pyserini toolkit as the evidence retriever and return top ten evidence corresponding to every claim.
\subsubsection{LLM based methods}
In this group, we select two novel models that extend and modify existing frameworks, ProgramFC\cite{ProgramFC} and FOLK\cite{FOLK}. ProgramFC is a baseline for verifying complex claims using LLMs for program decomposition. It contains both gold-evidence and open-book settings. We select N=1 for the setting of ProgramFC. FOLK is a recent baseline designed for an entirely open-book setting. So in the case of gold-evidence setting of FOLK, we will provide subclaims and corresponding gold evidence as input to the LLM, which then predicts the label.

\subsection{Main Results}

\begin{table*}[t!]
\centering
\caption{\textbf{Claim verification results on two datasets under gold and open settings. The accuracy are reported as the metrics. The best results are in bold respectively.}}
\label{tab:main results}
\begin{tabular*}{0.9\linewidth}{@{}llcccccccc@{}}
\toprule
\textbf{Models} & \textbf{} & \multicolumn{2}{c}{\textbf{HOVER (2-hop)}} & \multicolumn{2}{c}{\textbf{HOVER (3-hop)}} & \multicolumn{2}{c}{\textbf{HOVER (4-hop)}} & \multicolumn{2}{c}{\textbf{Feverous}} \\ 
\cmidrule(lr){3-4} \cmidrule(lr){5-6} \cmidrule(lr){7-8} \cmidrule(lr){9-10}
 &  & \textbf{Gold} & \textbf{Open} & \textbf{Gold} & \textbf{Open} & \textbf{Gold} & \textbf{Open} & \textbf{Gold} & \textbf{Open} \\ 
\midrule
\textbf{I} & BERT-FC (Soleimani et al., 2020) & 53.40 & 50.68 & 50.90 & 49.86 & 50.86 & 48.57 & 74.71 & 51.67 \\ 
 & LiST5 (Jiang et al., 2021) & 56.15 & 52.56 & 53.76 & 51.89 & 51.67 & 50.46 & 77.88 & 54.15 \\ 
 \midrule
\textbf{II} & RoBERTa-NLI (Nie et al., 2020) & 74.62 & 63.62 & 62.23 & 53.99 & 57.98 & 52.40 & 88.28 & 57.80 \\ 
 & DeBERTaV3-NLI (He et al., 2021) & 77.22 & 68.72 & 65.98 & 60.76 & 60.49 & 56.00 & 91.98 & 58.81 \\ 
 & MULTIVERS (Wadden et al., 2022b) & 68.86 & 60.17 & 59.87 & 52.55 & 55.67 & 51.86 & 86.03 & 56.61 \\ 
 \midrule
\textbf{III} & ProgramFC (Pan et al.,2023) & 74.10 & 69.36 & 66.13 & 60.63 & 65.69 & 59.16 & 91.77 & 67.80 \\
& FOLK (Wang et al.,2023) &73.69 & 66.26 & 64.64 & 54.80 & 63.10 & 60.35 & 89.93 & 67.01 \\
\midrule
\textbf{IV} & \textbf{CRAVE} & \textbf{77.89} & \textbf{69.80} & \textbf{75.90} & \textbf{69.63} & \textbf{72.86} & \textbf{65.83} & \textbf{93.55} & \textbf{74.95} \\
\bottomrule
\end{tabular*}
\end{table*}
We present the overall results for CRAVE compared to seven other baselines for claim verification in Table \ref{tab:main results}. CRAVE outperforms all baselines across all evaluation tasks, showcasing its strength and robustness, particularly in complex reasoning. Key observations include: 1) NLI fine-tuned models outperform pretrained models on various metrics, as they have been fine-tuned on multiple NLI tasks, improving their generalization. For instance, DeBERTaV3-NLI fine-tuned on Hover performs well on hover-related tasks. 2) ProgramFC and FOLK generally outperform the second category of models, especially in 4-hop reasoning, indicating that breaking down complex tasks into smaller sub-tasks is effective for solving complex reasoning problems. However, this advantage is less pronounced in 2-hop and 3-hop reasoning. 3) ProgramFC outperforms FOLK due to its open-book setting, which uses the Wikipedia corpus related to the datasets, whereas FOLK relies on the more general Google API. \par
In the full setting, integrating LLM reasoning with SLM judgment, CRAVE surpasses the best baseline in accuracy for both open-book and gold-evidence settings. Specifically, in the open-book setting, CRAVE achieves improvements of 0.44$\%$, 8.87$\%$, 5.48$\%$, and 7.15$\%$ on the Hover 2-hop, 3-hop, 4-hop and Feverous datasets respectively. In the gold-evidence setting, CRAVE improves accuracy by 0.67$\%$, 9.77$\%$, 7.17$\%$ and 1.57$\%$ on the same datasets. Observations include: 1) CRAVE demonstrates significant advantages in complex reasoning in both open-book and gold-evidence settings. 2) Compared to directly using LLMs for generating explanations and judgments (e.g., FOLK), CRAVE leverages LLMs to provide conflicting evidence—both supporting and refuting claims—while using the SLM judge. This approach significantly enhances accuracy by mitigating risks of hallucination and unsuitability of LLMs for directly handling complex reasoning tasks.

\subsection{Ablative Studies}
We perform ablative studies on several variants of CRAVE: 1) w/o Ambiguity Elimination (AE): Entities are retrieved from claims without resolving ambiguities; 2) w/o Entity Retrieval (ER): Claims are directly used to find evidence via BM25 after ambiguity elimination; 3) w/o Evidence Retrieval and Selection (ERS): Five sentences per entity are selected using BM25 without manual summaries; 4) w/o Conflicting Perspective Reasoning and Preliminary Judgment with LLM (LLM): Evidence and claims are used directly for SLM training and testing; 5) w/o judge module with SLM (SLM): LLM predictions are used as the final result. Some variants do not include a gold setting, as it provides evidence. \par

As shown in Table \ref{tab:ablation studies}, the ablative models experience varying degrees of performance degradation, highlighting the effectiveness of our proposed components. Specifically: 1) The accuracy drop without AE in the open setting results from unresolved subject ambiguities in complex claims, leading to incomplete evidence retrieval. As the number of hops increases, the accuracy degradation becomes more pronounced. 2) Without ER, using BM25 directly may lead to incomplete or inaccurate evidence retrieval, as BM25 often focuses on a single subject. 3) The absence of ERS leads to significant accuracy reduction, as summaries, which provide essential context, are omitted. 4) The accuracy drop without LLM in the open setting is due to noise in retrieved evidence. LLM can filter out through its semantic understanding, but SLMs struggle with. Meanwhile, LLM can assess the confidence of two rationales, aiding LLM to make final decesion. 5) The drop in accuracy without SLM confirms that directly using LLM outputs is inaccurate, particularly for complex claims, where even gold evidence performs poorly.

\begin{table}[t]
\centering
\caption{\textbf{Ablation studies by removing components from our proposed framework.}}
\label{tab:ablation studies}
\begin{footnotesize} 
\begin{tabular}{@{}l p{0.2cm} cccc@{}} 
\toprule
\textbf{Models} & \textbf{} & \textbf{HOVER (2)} & \textbf{HOVER (3)} & \textbf{HOVER (4)} & \textbf{Feverous} \\ 

\midrule
CRAVE & gold & 77.89 & 75.90 & 72.86 & 93.55 \\ 
 & open & 69.80 & 69.63 & 65.83 & 74.95  \\ 
 \midrule
w/o AE & gold & - & - & - & - \\ 
 & open & 68.92 & 68.32 & 64.64 & 74.81 \\ 
 \midrule
w/o ER & gold & - & - & - & -  \\
& open & 68.62 & 68.07 & 64.49 & 74.69 \\
\midrule
w/o ERS & gold & - & - & - & -  \\
& open & 67.50 & 67.63 & 63.52 & 65.96 \\
\midrule
w/o LLM &  gold & 70.43 & 68.09 & 70.27 & 92.64 \\
& open & 46.27 & 52.73 & 49.18 &47.73 \\
\midrule
w/o SLM &  gold & 73.09 & 62.65 & 59.10 & 82.53 \\
& open & 68.12 & 57.09 & 58.42 & 69.37 \\
\bottomrule
\end{tabular}
\end{footnotesize}
\end{table}

\subsection{Evaluation of Evidence Quality}

\begin{table}[t]
\centering
\caption{\textbf{Evaluation of evidence quality. The 'evidence' represents the ratio of retrieved gold evidence to the total gold evidence, and the 'claim' represents the ratio of claims for which at least one gold evidence is found.}}
\label{tab:explanation}
\begin{footnotesize} 
\begin{tabular}{@{}l p{0.3cm} cccc@{}} 
\toprule
\textbf{Models} & \textbf{} & \textbf{HOVER (2)} & \textbf{HOVER (3)} & \textbf{HOVER (4)} & \textbf{Feverous} \\ 
\midrule
CRAVE & evidence & 0.64 & 0.61 & 0.59 & 0.83 \\ 
 & claim & 0.83 & 0.89 & 0.92 & 0.86  \\ 
 \midrule
FOLK & evidence & 0.59 & 0.49 & 0.45 & 0.67 \\ 
 & claim & 0.83 & 0.87 & 0.92 & 0.80 \\ 
 \midrule
ProgramFC & evidence & 0.67 & 0.56 & 0.51 & 0.7  \\
& claim & 0.87 & 0.90 & 0.93 & 0.89 \\
\bottomrule
\end{tabular}
\end{footnotesize}
\end{table}

Since no existing metric evaluates the quality of retrieved evidence, we compare the ratio of gold evidence retrieved and the ratio of claims for which at least one gold evidence is found. The results is shown in Table \ref{tab:explanation}. With the addition of the ambiguity elimination and entity retrieval modules, our model outperforms others in gold evidence retrieval, except in the 2-hop Hover setting, where it slightly lags behind ProgramFC. After sentence decomposition, ProgramFC retrieves at least one gold evidence for most claims, but its retrieval is less comprehensive than CRAVE’s, especially for complex claims. CRAVE highlights a more thorough retrieval. Because LLM’s strong context and semantic understanding enable the model filter out redundancy and focus on key evidences when reasoning. This enables the SLM to make more accurate decisions based on the relevant clues identified by the LLM.

\subsection{The Impact of LLM Reasoning Aspects}\label{sec:4.5}
We explore the role of analytical perspectives for LLMs by testing the following scenarios: 1) The four perspectives used by CRAVE: Direct Evidence Analysis, Semantic Features and Relationships, Linguistic Patterns and Connections, and Logical Reasoning strictly from the evidence; 2) Removal of Direct Evidence Analysis; 3) Removal of Semantic Features and Relationships; 4) Removal of Logical Reasoning strictly from the evidence; 5) Removal of Linguistic Patterns and Connections; 6) No given perspective, allowing the LLM to generate explanations freely. The accuracy of the LLM judge is reported, as it directly influences the final model performance. \par

The experimental results in Table \ref{tab:reasoning aspects} show that: 1) Providing specific analytical perspectives improves the LLM's reasoning ability and accuracy, enabling it to extract relevant clues more effectively; 2) Removing any of the four perspectives causes varying degrees of accuracy decline, with the following importance hierarchy inferred: Direct Evidence Analysis $\textgreater$ Semantic Features and Relationships $\approx$ Linguistic Patterns and Connections $\textgreater$ Logical Reasoning strictly from the evidence. \par

The quality of LLM-generated explanations significantly affects the credibility of results, as they are presented to users for verification. A high-quality explanation should be comprehensive, logically consistent, readable, and credible. In this subsection, we use GPT-4 to evaluate explanation quality. We randomly selected 1,000 samples from the HOVER and FERVEROUS datasets and asked GPT-4 to score the explanations based on five criteria: 1) Conciseness, 2) Informativeness, 3) Persuasiveness, 4) Readability, and 5) Soundness, with scores from 1 (worst) to 5 (best). \par

\begin{figure*}[!t]
\centering
\subfloat[]{\includegraphics[width=2.5in]{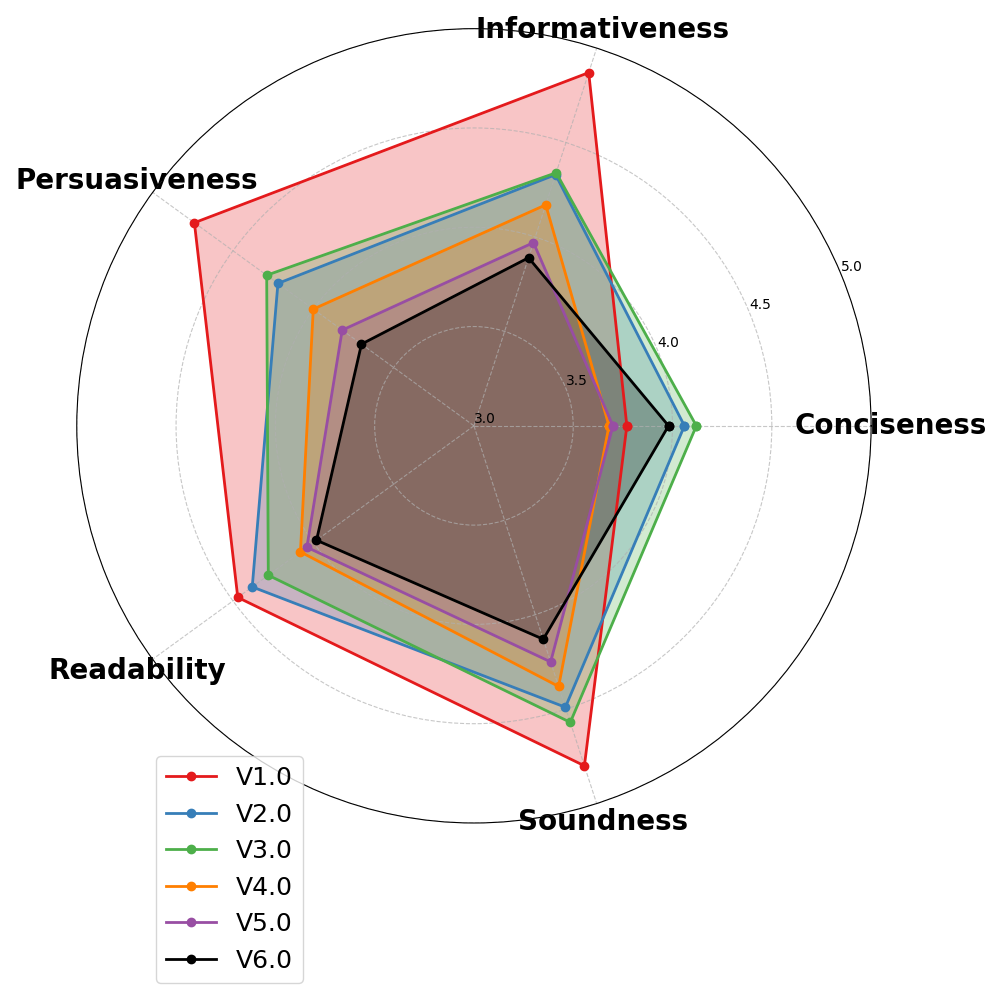}
\label{fig_first_case}}
\hfil
\subfloat[]{\includegraphics[width=2.5in]{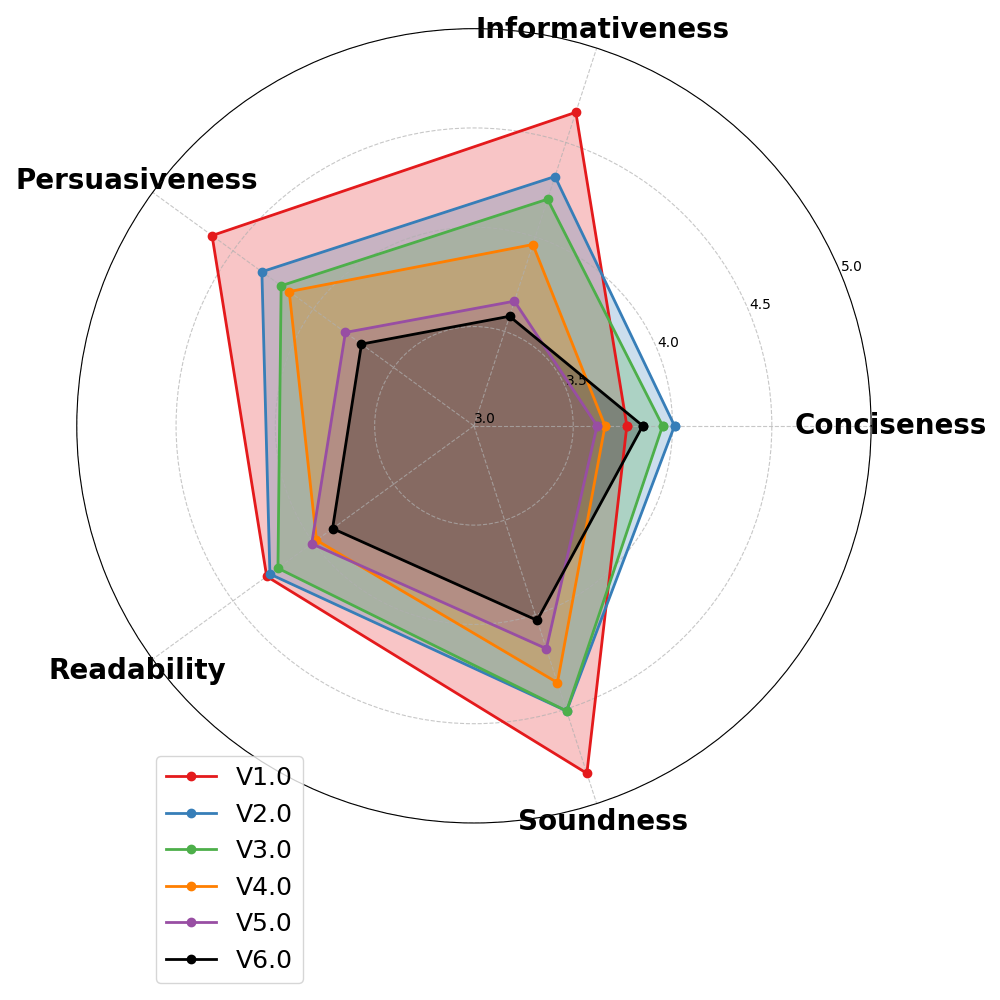}
\label{fig_second_case}}
\caption{Automatic GPT-4 evaluation of the explanation quality over different reasoning aspects. (a) Evaluation over HOVER dataset. (b) Evluation over FEVEROUS dataset}
\label{explanation}
\end{figure*}

The experimental results in Figure \ref{explanation} show: 1) CRAVE outperforms other methods, achieving the highest scores in all criteria except conciseness, demonstrating that the four perspectives enhance both accuracy and explanation quality; 2) Removing certain perspectives improves conciseness, indicating that some aspects introduce redundancy without clear support or refutation. The ordering of LLM reasoning angles based on these results allows the SLM judge module to focus on more important information through position embedding, leading to more accurate judgments.

\begin{table}[t]
\centering
\caption{\textbf{Accuracy over different versions of reasoning aspects. Detailed meaning of versions will be introduced in \ref{sec:4.5}}}
\label{tab:reasoning aspects}
\begin{footnotesize} 
\begin{tabular}{@{}lcccc@{}} 
\toprule
\textbf{Versions} & \textbf{HOVER (2)} & \textbf{HOVER (3)} & \textbf{HOVER (4)} & \textbf{Feverous} \\ 
\midrule
V1.0 & 68.12 & 57.09 & 58.42 & 69.37 \\ 
V2.0 & 66.07 & 58.35 & 55.49 & 67.65 \\ 
V3.0 & 65.01 & 56.19 & 57.80 & 69.31 \\ 
V4.0 & 66.61 & 58.43 & 58.53 & 68.99 \\ 
V5.0 & 64.56 & 56.29 & 58.71 & 68.87 \\ 
V6.0 & 64.27 & 56.14 & 55.32 & 66.65 \\ 
\bottomrule
\end{tabular}
\end{footnotesize}
\end{table}

\section{SLM input setting experiment}\label{sec:SLM settings}

\begin{table}[t]
\centering
\caption{\textbf{Accuracy over different SLM settings, exact meanings of different versions is explained in \ref{sec:SLM settings}}}
\label{tab:SLM setting}
\begin{footnotesize} 
\begin{tabular}{@{}lcccc@{}} 
\toprule
\textbf{Versions} & \textbf{HOVER (2)} & \textbf{HOVER (3)} & \textbf{HOVER (4)} & \textbf{Feverous} \\ 
\midrule
V1.0 & 74.87 & 74.48 & 72.09 & 92.38 \\ 
V2.0 & \textbf{77.89} & \textbf{75.90} & 72.86 & 93.55 \\ 
V3.0 & 77.62 & 75.19 & 71.90 & \textbf{93.65} \\ 
V4.0 & 76.29 & 74.86 & 72.09 & 93.20 \\ 
V5.0 & 72.29 & 69.74 & 67.37 & 93.04 \\ 
V6.0 & 76.02 & 75.08 & \textbf{73.82} & 92.90 \\ 
V7.0 & 77.80 & 74.65 & 72.86 & 93.31 \\
V8.0 & 77.09 & 73.12 & 69.30 & 93.04 \\
\bottomrule
\end{tabular}
\end{footnotesize}
\end{table}

This section discusses the approach used for setting the Small Language Model (SLM), experimenting with the following input processing methods (summarized in the table below): 1) RoBERTa with max truncation length (512); 2) RoBERTa with segment weights; 3) RoBERTa with self-attention, inputting only labels matching the LLM's predicted text; 4) RoBERTa with self-attention, inputting both claim and rationale texts; 5) GPT-2 with max truncation length (1024); 6) Fusing features of text1 and text2 before forming logits; 7) Splitting the text into claim, $rationale_true$, and $rationale_false$ without sequence fusion, then forming logits; 8) Similar to 7, but performing sequence fusion based on LLM’s output order before forming logits. \par

The specific experimental results are shown in Table \ref{tab:SLM setting}. To avoid interference from evidence quality, experiments were conducted with only gold evidence. The method using RoBERTa with segment weights, as used in CRAVE, achieved higher accuracy. It better integrates both supporting and refuting information compared to direct truncation and self-attention. Our approach also outperformed the feature fusion method, which fuses features before forming logits, likely due to the noise amplification caused by this step.

\section{Training method of SLM}

The training method for the SLM impacts its final accuracy. In this section, we compare different training and fine-tuning approaches, evaluating the model on two datasets in both gold and open settings. The experimental results are summarized in Table \ref{tab:train methods}. \par

Our goal is for the model to perform well in both gold and open settings, minimizing dependency on gold evidence while demonstrating strong generalization and robustness. Based on the experimental data, we draw the following conclusions: 1) Models trained on gold evidence perform poorly in the open setting, while those trained in the open setting show some generalization ability when evaluated on gold evidence, suggesting greater robustness. 2) Fine-tuning models trained on gold evidence with different data ratios shows that accuracy increases initially and then decreases as the open setting ratio rises. The optimal performance occurs at an 80:20 open:gold ratio. This indicates that adjusting the data ratio can significantly improve accuracy in different settings. \par

In conclusion, careful selection of training methods and data ratio adjustment can improve the model’s accuracy and adaptability to diverse data settings.

\begin{table*}[t]
\centering
\caption{\textbf{Evaluation of different SLM training method. CFT means continue fine-tuning and the digits in parentheses are data ratio. The accuracy are reported as the metrics.}}
\label{tab:train methods}
\begin{tabular*}{0.9\linewidth}{@{}llcccccccc@{}}
\toprule
\textbf{Models} & \textbf{} & \multicolumn{2}{c}{\textbf{HOVER (2-hop)}} & \multicolumn{2}{c}{\textbf{HOVER (3-hop)}} & \multicolumn{2}{c}{\textbf{HOVER (4-hop)}} & \multicolumn{2}{c}{\textbf{Feverous}} \\ 
\cmidrule(lr){3-4} \cmidrule(lr){5-6} \cmidrule(lr){7-8} \cmidrule(lr){9-10}
 &  & \textbf{Gold} & \textbf{Open} & \textbf{Gold} & \textbf{Open} & \textbf{Gold} & \textbf{Open} & \textbf{Gold} & \textbf{Open} \\ 
\midrule
\textbf{I} & Training on gold data & 77.89 & 64.56 & 75.90 & 58.40 & 72.86 & 57.07 & 93.55 & 66.63 \\
& Training on open data & 72.02 & 69.80 & 73.39 & 69.63 & 72.18 & 65.83 & 88.58 & 74.95 \\
& Training on mix data(open:gold=1:1) & 75.49 & 68.03 & 71.43 & 66.14 & 68.14 & 64.10 & 92.43 & 72.48 \\
 \midrule
\textbf{II} & training on gold+CFT(open:gold=100:0) & 69.80&64.92& 70.83 & 68.97 & 70.93 & 66.51 & 89.22 & 84.68 \\
& training on gold+CFT(open:gold=95:5) & 66.25 & 65.72 & 66.69 & 66.36 & 64.68 & 66.31 & 88.75 &73.56 \\
& training on gold+CFT(open:gold=90:10) & 71.05 & 67.23 & 72.08 & 68.43 & 71.90 & 66.51 & 89.35 & 74.07 \\
& training on gold+CFT(open:gold=85:15) & 71.05 & 67.14 & 72.14 & 69.08 & 69.68 & 63.14 & 90.98 & 74.21 \\
& training on gold+CFT(open:gold=80:20) & 75.58 & 69.36 & 74.32 & 67.67 & 71.41 & 62.58 & 89.79 & 74.31 \\
& training on gold+CFT(open:gold=75:25) & 72.38 & 66.25& 72.30 & 67.28 & 71.13 & 65.05 & 90.06 & 72.58 \\
\bottomrule
\end{tabular*}
\end{table*}

\section{CONCLUSION AND FUTURE WORK}
We propose CRAVE, an explainable model for claim verification. The CRAVE model performs ambiguity elimination on the input claim and retrieves evidence based on entities. Then, leveraging the strong semantic understanding and reasoning capabilities of LLMs, it generates clues indicating whether the evidence supports or refutes the claim and make a preliminary judgment assessing the confidence of both clues. Finally, it use SLM to map the encoded inputs of the claim and clues to a label output. CRAVE effectively integrates the reasoning capabilities of LLMs with the fitting accuracy of SLMs and utilizes four reasoning aspects to enhance the inference performance of LLMs. CRAVE achieves state-of-the-art results on both the HOVER and FEVEROUS datasets, demonstrating superior evidence extraction capabilities and explanation quality. \par 
In future work, we plan to explore the impact of LLM scale on reasoning ability and extend the detection tasks from textual claims to unstructured claim detection.

\section*{Acknowledgments}
This work was supported by the National Key Research and Development Plan in China (No. 2023YFC3306100), National Natural Science Foundation of China (No. 62172278, 62002219).

\bibliographystyle{IEEEtranS}

\begin{thebibliography}{10}
\providecommand{\url}[1]{#1}
\csname url@samestyle\endcsname
\providecommand{\newblock}{\relax}
\providecommand{\bibinfo}[2]{#2}
\providecommand{\BIBentrySTDinterwordspacing}{\spaceskip=0pt\relax}
\providecommand{\BIBentryALTinterwordstretchfactor}{4}
\providecommand{\BIBentryALTinterwordspacing}{\spaceskip=\fontdimen2\font plus
\BIBentryALTinterwordstretchfactor\fontdimen3\font minus \fontdimen4\font\relax}
\providecommand{\BIBforeignlanguage}[2]{{%
\expandafter\ifx\csname l@#1\endcsname\relax
\typeout{** WARNING: IEEEtranS.bst: No hyphenation pattern has been}%
\typeout{** loaded for the language `#1'. Using the pattern for}%
\typeout{** the default language instead.}%
\else
\language=\csname l@#1\endcsname
\fi
#2}}
\providecommand{\BIBdecl}{\relax}
\BIBdecl

\bibitem{neturalnetwork}
\BIBentryALTinterwordspacing
J.~Alghamdi, Y.~Lin, and S.~Luo, ``Towards covid-19 fake news detection using transformer-based models,'' \emph{Knowledge-Based Systems}, vol. 274, pp. 110\,642 -- 110\,642, 2023. [Online]. Available: \url{https://api.semanticscholar.org/CorpusID:258807287}
\BIBentrySTDinterwordspacing

\bibitem{feverous}
\BIBentryALTinterwordspacing
R.~Aly, Z.~Guo, M.~Schlichtkrull, J.~Thorne, A.~Vlachos, C.~Christodoulopoulos, O.~Cocarascu, and A.~Mittal, ``Feverous: Fact extraction and verification over unstructured and structured information,'' 2021. [Online]. Available: \url{https://arxiv.org/abs/2106.05707}
\BIBentrySTDinterwordspacing

\bibitem{hallucinate1}
\BIBentryALTinterwordspacing
Y.~Bang, S.~Cahyawijaya, N.~Lee, W.~Dai, D.~Su, B.~Wilie, H.~Lovenia, Z.~Ji, T.~Yu, W.~Chung, Q.~V. Do, Y.~Xu, and P.~Fung, ``A multitask, multilingual, multimodal evaluation of chatgpt on reasoning, hallucination, and interactivity,'' 2023. [Online]. Available: \url{https://arxiv.org/abs/2302.04023}
\BIBentrySTDinterwordspacing

\bibitem{ambi}
\BIBentryALTinterwordspacing
M.~Glockner, I.~Stali{\={u}}nait{\.{e}}, J.~Thorne, G.~Vallejo, A.~Vlachos, and I.~Gurevych, ``{A}mbi{FC}: Fact-checking ambiguous claims with evidence,'' \emph{Transactions of the Association for Computational Linguistics}, vol.~12, pp. 1--18, 2024. [Online]. Available: \url{https://aclanthology.org/2024.tacl-1.1/}
\BIBentrySTDinterwordspacing

\bibitem{LLMpotential3}
\BIBentryALTinterwordspacing
J.~Guan, J.~Dodge, D.~Wadden, M.~Huang, and H.~Peng, ``Language models hallucinate, but may excel at fact verification,'' 2024. [Online]. Available: \url{https://arxiv.org/abs/2310.14564}
\BIBentrySTDinterwordspacing

\bibitem{automatedfactchecking}
\BIBentryALTinterwordspacing
Z.~Guo, M.~Schlichtkrull, and A.~Vlachos, ``A survey on automated fact-checking,'' \emph{Transactions of the Association for Computational Linguistics}, vol.~10, pp. 178--206, 2022. [Online]. Available: \url{https://aclanthology.org/2022.tacl-1.11/}
\BIBentrySTDinterwordspacing

\bibitem{Deberta}
\BIBentryALTinterwordspacing
P.~He, J.~Gao, and W.~Chen, ``Debertav3: Improving deberta using electra-style pre-training with gradient-disentangled embedding sharing,'' 2023. [Online]. Available: \url{https://arxiv.org/abs/2111.09543}
\BIBentrySTDinterwordspacing

\bibitem{GoodActor}
\BIBentryALTinterwordspacing
B.~Hu, Q.~Sheng, J.~Cao, Y.~Shi, Y.~Li, D.~Wang, and P.~Qi, ``Bad actor, good advisor: Exploring the role of large language models in fake news detection,'' \emph{Proceedings of the AAAI Conference on Artificial Intelligence}, vol.~38, no.~20, p. 22105–22113, Mar. 2024. [Online]. Available: \url{http://dx.doi.org/10.1609/aaai.v38i20.30214}
\BIBentrySTDinterwordspacing

\bibitem{LLMdirect2}
\BIBentryALTinterwordspacing
Y.~Huang and L.~Sun, ``Fakegpt: Fake news generation, explanation and detection of large language models,'' 2024. [Online]. Available: \url{https://arxiv.org/abs/2310.05046}
\BIBentrySTDinterwordspacing

\bibitem{list5}
\BIBentryALTinterwordspacing
K.~Jiang, R.~Pradeep, and J.~Lin, ``Exploring listwise evidence reasoning with t5 for fact verification,'' in \emph{Proceedings of the 59th Annual Meeting of the Association for Computational Linguistics and the 11th International Joint Conference on Natural Language Processing (Volume 2: Short Papers)}, C.~Zong, F.~Xia, W.~Li, and R.~Navigli, Eds.\hskip 1em plus 0.5em minus 0.4em\relax Online: Association for Computational Linguistics, Aug. 2021, pp. 402--410. [Online]. Available: \url{https://aclanthology.org/2021.acl-short.51/}
\BIBentrySTDinterwordspacing

\bibitem{Hover}
\BIBentryALTinterwordspacing
Y.~Jiang, S.~Bordia, Z.~Zhong, C.~Dognin, M.~Singh, and M.~Bansal, ``{H}o{V}er: A dataset for many-hop fact extraction and claim verification,'' in \emph{Findings of the Association for Computational Linguistics: EMNLP 2020}, T.~Cohn, Y.~He, and Y.~Liu, Eds.\hskip 1em plus 0.5em minus 0.4em\relax Online: Association for Computational Linguistics, Nov. 2020, pp. 3441--3460. [Online]. Available: \url{https://aclanthology.org/2020.findings-emnlp.309/}
\BIBentrySTDinterwordspacing

\bibitem{damage2}
\BIBentryALTinterwordspacing
Y.~Jin, Y.-C. Lee, K.~Sharma, M.~Ye, K.~Sikka, A.~Divakaran, and S.~Kumar, ``Predicting information pathways across online communities,'' in \emph{Proceedings of the 29th ACM SIGKDD Conference on Knowledge Discovery and Data Mining}, ser. KDD '23.\hskip 1em plus 0.5em minus 0.4em\relax New York, NY, USA: Association for Computing Machinery, 2023, p. 1044–1056. [Online]. Available: \url{https://doi.org/10.1145/3580305.3599470}
\BIBentrySTDinterwordspacing

\bibitem{FactKG}
\BIBentryALTinterwordspacing
J.~Kim, S.~Park, Y.~Kwon, Y.~Jo, J.~Thorne, and E.~Choi, ``{F}act{KG}: Fact verification via reasoning on knowledge graphs,'' in \emph{Proceedings of the 61st Annual Meeting of the Association for Computational Linguistics (Volume 1: Long Papers)}, A.~Rogers, J.~Boyd-Graber, and N.~Okazaki, Eds.\hskip 1em plus 0.5em minus 0.4em\relax Toronto, Canada: Association for Computational Linguistics, Jul. 2023, pp. 16\,190--16\,206. [Online]. Available: \url{https://aclanthology.org/2023.acl-long.895/}
\BIBentrySTDinterwordspacing

\bibitem{LLMdebate2}
\BIBentryALTinterwordspacing
K.~Kim, S.~Lee, K.-H. Huang, H.~P. Chan, M.~Li, and H.~Ji, ``Can llms produce faithful explanations for fact-checking? towards faithful explainable fact-checking via multi-agent debate,'' 2024. [Online]. Available: \url{https://arxiv.org/abs/2402.07401}
\BIBentrySTDinterwordspacing

\bibitem{MUSER}
\BIBentryALTinterwordspacing
H.~Liao, J.~Peng, Z.~Huang, W.~Zhang, G.~Li, K.~Shu, and X.~Xie, ``Muser: A multi-step evidence retrieval enhancement framework for fake news detection,'' \emph{Proceedings of the 29th ACM SIGKDD Conference on Knowledge Discovery and Data Mining}, 2023. [Online]. Available: \url{https://api.semanticscholar.org/CorpusID:259244038}
\BIBentrySTDinterwordspacing

\bibitem{Harmfulmeme}
\BIBentryALTinterwordspacing
H.~Lin, Z.~Luo, W.~Gao, J.~Ma, B.~Wang, and R.~Yang, ``Towards explainable harmful meme detection through multimodal debate between large language models,'' \emph{Proceedings of the ACM on Web Conference 2024}, 2024. [Online]. Available: \url{https://api.semanticscholar.org/CorpusID:267199823}
\BIBentrySTDinterwordspacing

\bibitem{Roberta-NLI}
\BIBentryALTinterwordspacing
Y.~Nie, A.~Williams, E.~Dinan, M.~Bansal, J.~Weston, and D.~Kiela, ``Adversarial nli: A new benchmark for natural language understanding,'' 2020. [Online]. Available: \url{https://arxiv.org/abs/1910.14599}
\BIBentrySTDinterwordspacing

\bibitem{ProgramFC}
\BIBentryALTinterwordspacing
L.~Pan, X.~Wu, X.~Lu, A.~T. Luu, W.~Y. Wang, M.-Y. Kan, and P.~Nakov, ``Fact-checking complex claims with program-guided reasoning,'' in \emph{Proceedings of the 61st Annual Meeting of the Association for Computational Linguistics (Volume 1: Long Papers)}, A.~Rogers, J.~Boyd-Graber, and N.~Okazaki, Eds.\hskip 1em plus 0.5em minus 0.4em\relax Toronto, Canada: Association for Computational Linguistics, Jul. 2023, pp. 6981--7004. [Online]. Available: \url{https://aclanthology.org/2023.acl-long.386/}
\BIBentrySTDinterwordspacing

\bibitem{5WQA}
\BIBentryALTinterwordspacing
A.~Rani, S.~T.~I. Tonmoy, D.~Dalal, S.~Gautam, M.~Chakraborty, A.~Chadha, A.~Sheth, and A.~Das, ``{FACTIFY}-5{WQA}: 5{W} aspect-based fact verification through question answering,'' in \emph{Proceedings of the 61st Annual Meeting of the Association for Computational Linguistics (Volume 1: Long Papers)}, A.~Rogers, J.~Boyd-Graber, and N.~Okazaki, Eds.\hskip 1em plus 0.5em minus 0.4em\relax Toronto, Canada: Association for Computational Linguistics, Jul. 2023, pp. 10\,421--10\,440. [Online]. Available: \url{https://aclanthology.org/2023.acl-long.581/}
\BIBentrySTDinterwordspacing

\bibitem{fouraspects}
\BIBentryALTinterwordspacing
C.~Scheibenzuber, L.-M. Neagu, S.~Ruseti, B.~Artmann, C.~Bartsch, M.~Kubik, M.~Dascalu, S.~Trausan-Matu, and N.~Nistor, ``Dialog in the echo chamber: Fake news framing predicts emotion, argumentation and dialogic social knowledge building in subsequent online discussions,'' \emph{Computers in Human Behavior}, vol. 140, p. 107587, 2023. [Online]. Available: \url{https://www.sciencedirect.com/science/article/pii/S0747563222004071}
\BIBentrySTDinterwordspacing

\bibitem{BERT-FC}
\BIBentryALTinterwordspacing
A.~Soleimani, C.~Monz, and M.~Worring, ``Bert for evidence retrieval and claim verification,'' 2019. [Online]. Available: \url{https://arxiv.org/abs/1910.02655}
\BIBentrySTDinterwordspacing

\bibitem{pipeline4}
\BIBentryALTinterwordspacing
D.~Wadden, K.~Lo, L.~L. Wang, A.~Cohan, I.~Beltagy, and H.~Hajishirzi, ``{M}ulti{V}er{S}: Improving scientific claim verification with weak supervision and full-document context,'' in \emph{Findings of the Association for Computational Linguistics: NAACL 2022}, M.~Carpuat, M.-C. de~Marneffe, and I.~V. Meza~Ruiz, Eds.\hskip 1em plus 0.5em minus 0.4em\relax Seattle, United States: Association for Computational Linguistics, Jul. 2022, pp. 61--76. [Online]. Available: \url{https://aclanthology.org/2022.findings-naacl.6/}
\BIBentrySTDinterwordspacing

\bibitem{multivers}
\BIBentryALTinterwordspacing
D.~Wadden, K.~Lo, and L.~L. e.~a. Wang, ``{M}ulti{V}er{S}: Improving scientific claim verification with weak supervision and full-document context,'' in \emph{Findings of the Association for Computational Linguistics: NAACL 2022}, M.~Carpuat, M.-C. de~Marneffe, and I.~V. Meza~Ruiz, Eds.\hskip 1em plus 0.5em minus 0.4em\relax Seattle, United States: Association for Computational Linguistics, Jul. 2022, pp. 61--76. [Online]. Available: \url{https://aclanthology.org/2022.findings-naacl.6/}
\BIBentrySTDinterwordspacing

\bibitem{LLMpotential2}
\BIBentryALTinterwordspacing
H.~Wan, S.~Feng, Z.~Tan, H.~Wang, Y.~Tsvetkov, and M.~Luo, ``Dell: Generating reactions and explanations for llm-based misinformation detection,'' 2024. [Online]. Available: \url{https://arxiv.org/abs/2402.10426}
\BIBentrySTDinterwordspacing

\bibitem{LLMdebate1}
\BIBentryALTinterwordspacing
B.~Wang, J.~Ma, H.~Lin, Z.~Yang, R.~Yang, Y.~Tian, and Y.~Chang, ``Explainable fake news detection with large language model via defense among competing wisdom,'' 2024. [Online]. Available: \url{https://arxiv.org/abs/2405.03371}
\BIBentrySTDinterwordspacing

\bibitem{FOLK}
\BIBentryALTinterwordspacing
H.~Wang and K.~Shu, ``Explainable claim verification via knowledge-grounded reasoning with large language models,'' in \emph{Findings of the Association for Computational Linguistics: EMNLP 2023}, H.~Bouamor, J.~Pino, and K.~Bali, Eds.\hskip 1em plus 0.5em minus 0.4em\relax Singapore: Association for Computational Linguistics, Dec. 2023, pp. 6288--6304. [Online]. Available: \url{https://aclanthology.org/2023.findings-emnlp.416/}
\BIBentrySTDinterwordspacing

\bibitem{hallucinate3}
\BIBentryALTinterwordspacing
Y.~Zhang, Y.~Li, L.~Cui, D.~Cai, L.~Liu, T.~Fu, X.~Huang, E.~Zhao, Y.~Zhang, Y.~Chen, L.~Wang, A.~T. Luu, W.~Bi, F.~Shi, and S.~Shi, ``Siren's song in the ai ocean: A survey on hallucination in large language models,'' 2023. [Online]. Available: \url{https://arxiv.org/abs/2309.01219}
\BIBentrySTDinterwordspacing

\bibitem{fouraspect2}
\BIBentryALTinterwordspacing
C.~Zhou, K.~Li, and Y.~Lu, ``Linguistic characteristics and the dissemination of misinformation in social media: The moderating effect of information richness,'' \emph{Information Processing \& Management}, vol.~58, no.~6, p. 102679, 2021. [Online]. Available: \url{https://www.sciencedirect.com/science/article/pii/S0306457321001655}
\BIBentrySTDinterwordspacing

\end{thebibliography}

\end{document}